\documentclass[letterpaper,10pt,conference]{ieeeconf}
\IEEEoverridecommandlockouts
\usepackage{amsmath,amssymb,booktabs,cite,graphicx,xcolor,url,multirow}
\graphicspath{{figs/}}

\title{\LARGE \bf
From Prediction to Decision: World-Model-Guided\\
Action Selection for Continuous Pile Excavation}
\author{%
Ailing Zhang$^{1,2}$, Fan Gao$^{1}$, Song Zhang$^{1,*}$,\\
Kawa Leong$^{1,2}$, Ziyu Wu$^{1,3}$, Yafei Wang$^{2}$%
\thanks{$^{1}$Tsing-AI (Shanghai) Technology Co., Ltd.}%
\thanks{$^{2}$Shanghai Jiao Tong University.}%
\thanks{$^{3}$Shanghai Ocean University.}%
\thanks{$^{*}$Corresponding author: \mbox{Song Zhang},
\textit{song.zhang@zex-t.com}.}%
}

\begin{document}
\maketitle
\thispagestyle{empty}
\pagestyle{empty}

\begin{abstract}

Wheel-loader excavation is a sequential decision problem in which every scoop changes the terrain available to subsequent actions. A practical world model must predict action consequences accurately, rank candidates in real time, and operate inside the closed loop of a full-size machine. We present the \emph{World-Action Model} (WAM), which proposes multiple scoops, rejects geometrically inadmissible candidates, jointly predicts signed terrain change and loaded volume, executes the candidate with the largest predicted load, and replans from the newly observed terrain. On 32 geometry-disjoint MinSlope test episodes, adding world-model ranking to matched diffusion proposals reduces the mean scoop count from 651.8 to 540.6 (17.1\%), preserves 32/32 completion, and improves every paired episode. In a complete-system comparison, WAM completes 32/32 episodes versus 29/32 for an independently trained soft actor--critic policy. Comparisons of input representations, spatial support, and five architectures identify an accurate and efficient physics-structured predictor. We further evaluate the interface on event-disjoint full-size-loader data and deploy the complete perception--proposal--prediction--selection--execution loop for autonomous excavation. The ROS2/TensorRT implementation processes five candidates in 72.4\,ms on a Jetson AGX Orin. The simulation results establish decision-level gains, while the physical experiments demonstrate real-world closed-loop feasibility.

\end{abstract}

\section{Introduction}
Wheel-loader excavation modifies the terrain on which subsequent actions operate. Each scoop removes and redistributes material, determining the immediate load and the future working face. Because the resulting geometry can preserve or degrade subsequent excavation conditions, decisions remain coupled throughout continuous depletion. Action-conditioned prediction can therefore evaluate alternatives before an irreversible terrain modification. Existing earthmoving systems integrate perception, planning, and hydraulic control~\cite{zhang2021autonomous}, while learned controllers adapt digging motions to soil conditions~\cite{egli2022soil,franceschini2025oscillatory}. Continuous depletion further requires repeatedly selecting an admissible scoop as the machine reshapes its environment.

World models address this need by predicting how candidate actions change the physical scene and what loading outcomes they produce. Candidate evaluation through learned dynamics has improved control and manipulation~\cite{williams2017mppi,nagabandi2018neural,hansen2024tdmpc2,finn2017foresight}. Prior wheel-loader systems establish two complementary routes: world models predict loading outcomes and support finite-horizon planning in simulation~\cite{aoshima2024world,aoshima2025endtoend}, while a learned dream environment can train a policy that is subsequently deployed on a full-size loader~\cite{eriksson2024dream}. Less explored is deploying the world model itself as an execution-time decision module: predicting the consequences of multiple candidates onboard before each physical scoop, selecting one action, and updating the decision from the observed result. We study this role at two levels, isolating the value of consequence-based ranking during geometry-disjoint near-complete depletion and deploying the same prediction--selection loop for autonomous excavation on a full-size loader.

Our World-Action Model (WAM) addresses this question through multimodal diffusion proposals, an independent geometric admissibility filter, and one-step world-model ranking. Each decision uses only the current observation and predicted outcomes of the current candidates. WAM then executes one selected action, observes the resulting terrain, and repeats the same decision cycle until the depletion criterion is reached.

\begin{figure*}[t]
\centering
\includegraphics[width=\textwidth]{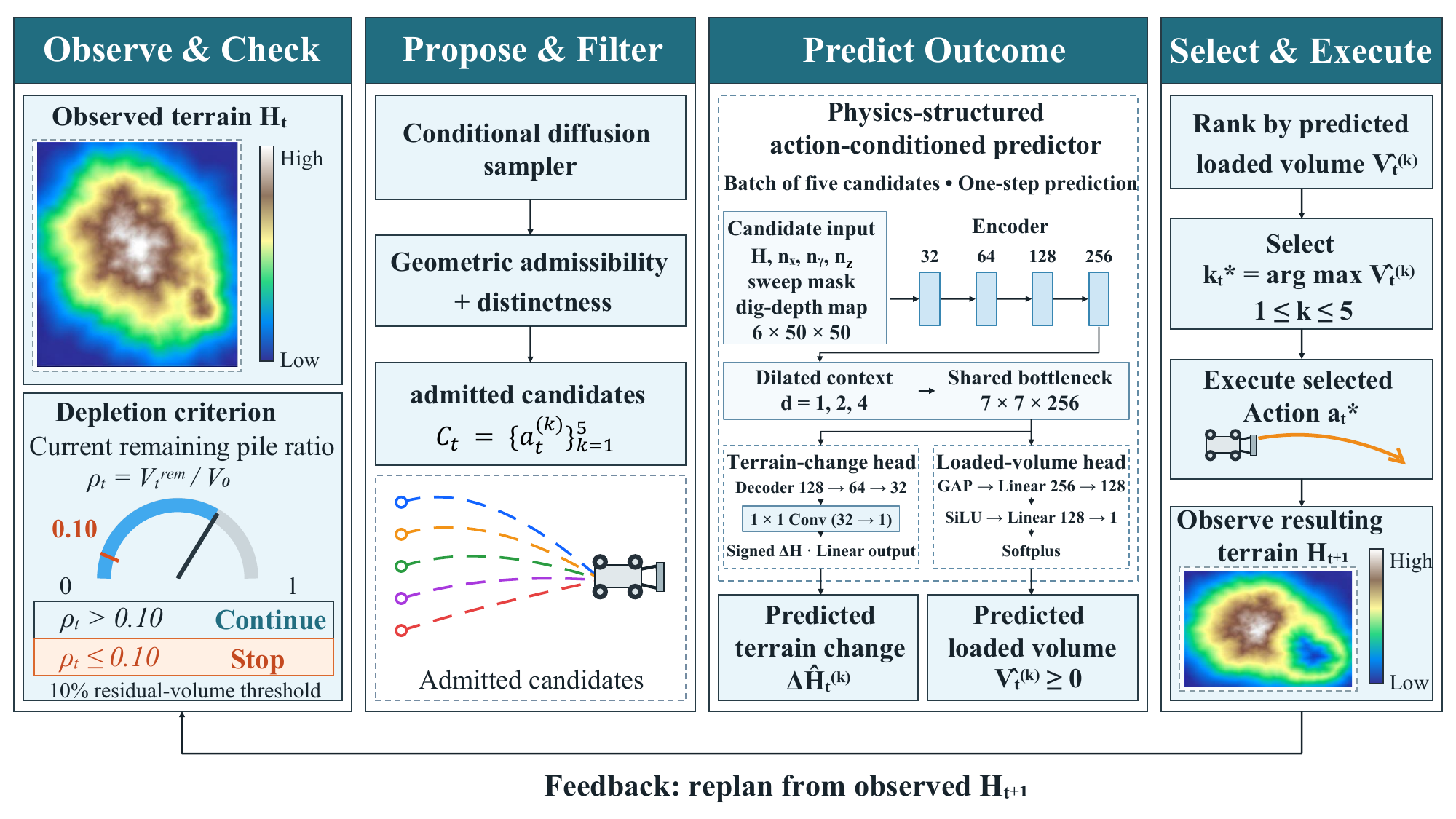}
\caption{Overview of WAM for continuous pile depletion. Given the observed terrain, a conditional diffusion model proposes scoop actions, which are filtered for geometric admissibility and distinctness. A frozen action-conditioned world model jointly predicts signed terrain change and loaded volume for the admitted candidates. WAM executes the candidate with the highest predicted loaded volume and replans using the newly observed terrain. The cycle repeats until the remaining pile volume is at most 10\% of its initial value.}
\label{fig:overview}
\end{figure*}

Figure~\ref{fig:overview} summarizes WAM and its simulation-to-deployment evaluation. The contributions are threefold.
\begin{itemize}
\setlength{\itemsep}{1pt}\setlength{\topsep}{2pt}\setlength{\parsep}{0pt}
\raggedright
\item \textbf{Proposal–selection WAM:} We introduce a closed-loop decision process that generates multimodal scoop candidates with diffusion, filters them for geometric admissibility, evaluates them in batch using a frozen action-conditioned world model, executes the highest-ranked action, and replans from the newly observed terrain.
\item \textbf{Physics-structured action-conditioned predictor:} We develop a pre-execution predictor that represents each candidate using terrain height, surface-normal channels, and trajectory-derived sweep and dig-depth maps to jointly predict signed terrain change and loaded volume. 
\item \textbf{Real-world closed-loop deployment:} We deploy the full perception-proposal-prediction-selection-execution loop onboard a full-size wheel loader for autonomous excavation. The ROS2/TensorRT implementation evaluates five candidates in 72.4\,ms on a Jetson AGX Orin, establishing real-machine integration and closed-loop feasibility.
\end{itemize}
MinSlope provides the quantitative decision comparison, whereas the physical experiments establish real-machine integration and closed-loop operation rather than a statistically validated real-world efficiency gain.

\section{Related Work}
\subsection{Terrain Dynamics and World Models}
World models predict future states or outcomes for planning~\cite{ha2018worldmodels,hafner2020dreamer}. Aoshima \emph{et al.} learn wheel-loader pile-state and loading models for finite-horizon tree search over simulated loading sequences~\cite{aoshima2024world,aoshima2025endtoend}. Eriksson \emph{et al.} train an RL bucket-filling policy in a dream environment learned from recorded loading cycles and deploy it on a full-size loader~\cite{eriksson2024dream}. L-GBND localizes graph prediction for terrain manipulation, while graph dynamics also represent deformable configurations~\cite{liu2025localized,deng2024graph}. WAM instead keeps its frozen world model onboard to directly rank multiple admissible scoops before physical execution and replan from the newly observed terrain. Matched-candidate simulation comparisons isolate the ranking effect through geometry-disjoint depletion to 10\% residual volume, while full-size-loader trials verify closed-loop integration.

Granular response depends on material properties, which may be inferred for simulator calibration~\cite{matl2020granular}. Real-time and multiscale terrain models trade fidelity for throughput~\cite{millard2023granulargym,servin2021multiscale}; Chrono, Isaac Sim, and Newton provide complementary DEM, robot-learning, and MPM pipelines~\cite{fang2021chronogpu,nvidia2026isaacsim,newton2025}. We use these pipelines to test whether the same data interface can be learned independently in each domain.

\subsection{Excavation Policies and Admissible Action Selection}
Learning-based excavation addresses trajectory generation and soil adaptation~\cite{lu2022geometric,egli2022soil}. Deep-RL wheel-loader controllers predict lift, tilt, vehicle velocity, or dig-point commands~\cite{azulay2021wheel,backman2021continuous}, and data-efficient adaptation has been demonstrated on a full-size loader~\cite{eriksson2024unknown}. These studies motivate SAC as the model-free baseline. Complementary work improves jamming resistance with oscillatory primitives~\cite{franceschini2025oscillatory} and separates learned behavior generation from verifiable execution constraints~\cite{benton2024verifiable}. WAM similarly assigns proposal, admissibility, and ranking to separate modules: diffusion supplies multimodal actions~\cite{chi2024diffusion}, geometry rejects infeasible actions, and the world model selects among the remainder.

\section{Methodology}
\subsection{Problem Formulation and Action Representation}
Let $H_t\in\mathbb{R}^{W\times W}$ denote the pre-action height map, $\mathbf n_t\in\mathbb{R}^{3\times W\times W}$ its surface-normal channels, and $\rho_t$ the ratio of remaining to initial pile volume. WAM produces the normalized action
\begin{equation}
\mathbf a_t=
[\psi_t,y_t,d_t^{\rm app},\ell_t,d_t^{\max},v_t,\kappa_t]^\top
\in[-1,1]^7,
\label{eq:action}
\end{equation}

The seven coordinates represent approach heading, lateral work-face
offset, approach distance, cutting length, maximum cutting depth,
speed scale, and bucket-curl angle, respectively. In the MinSlope
experiments, each normalized coordinate $a_{t,j}\in[-1,1]$ is mapped
to its physical value through
\begin{equation}
q_{t,j}=q_j^{\min}
+\frac{a_{t,j}+1}{2}
\left(q_j^{\max}-q_j^{\min}\right).
\label{eq:action-mapping}
\end{equation}
The MinSlope-specific physical limits shared by WAM and SAC are
listed in Table~\ref{tab:action-range}.

\begin{table}[t]
\caption{Physical ranges of the seven-dimensional excavation action.}
\label{tab:action-range}
\centering
\setlength{\tabcolsep}{3.2pt}
\begin{tabular}{@{}lll@{}}
\toprule
Coordinate & Physical variable & Range\\
\midrule
$\psi_t$ & Approach heading & $[165^\circ,195^\circ]$\\
$y_t$ & Lateral offset & $[-1.25,1.25]\,\mathrm m$\\
$d_t^{\rm app}$ & Approach distance & $[3.0,5.0]\,\mathrm m$\\
$\ell_t$ & Cutting length & $[2.5,4.5]\,\mathrm m$\\
$d_t^{\max}$ & Maximum cutting depth & $[0.20,0.55]\,\mathrm m$\\
$v_t$ & Speed scale & $[0.70,1.15]$\\
$\kappa_t$ & Bucket-curl angle & $[44^\circ,52^\circ]$\\
\bottomrule
\end{tabular}
\end{table}

In the MinSlope environment, a deterministic trajectory generator
converts the physical action into a bucket path, and the bucket
capacity is $C=6\,\mathrm{m^3}$. Pitch, roll, and lift height are
determined by the generated trajectory rather than specified as
independent action coordinates.

We rasterize the resulting path into a swept-area mask $m_t$ and dig-depth map $d_t$. The world model predicts
\begin{equation}
(\Delta\hat H_t,\hat V_t)=f_\theta(H_t,\mathbf n_t,m_t,d_t),
\label{eq:wm}
\end{equation}
where $\Delta H_t=H_{t+1}-H_t$ is signed terrain change and $V_t$ is loaded volume. Only information available before executing $\mathbf a_t$ enters~(\ref{eq:wm}). After observing $H_{t+1}$, the process repeats.

\subsection{Joint Terrain--Volume World Model}
For each candidate action, the predictor operates on a $10\,\mathrm m\times10\,\mathrm m$ region of interest discretized into $50\times50$ cells. Its six-channel input comprises the current height map, three surface-normal components, the swept-area mask, and the dig-depth map. Input channels are zero-centered and normalized channel-wise using domain-specific physical scales. The terrain-change and loaded-volume targets are normalized as
\begin{equation}
\widetilde{\Delta H}=\frac{\Delta H}{s_H},
\qquad
\widetilde V=\frac{V}{s_V},
\label{eq:wm-normalization}
\end{equation}
where \(s_H>0\) and \(s_V>0\) are domain-specific normalization scales for terrain change and loaded volume, respectively. For each simulation or real-loader domain, \(s_H\) is set to the 99th percentile of \(|\Delta H|\) over valid pixels in the corresponding training split, and \(s_V\) is set to the calibrated bucket capacity when available. All data-derived normalization statistics are computed exclusively from the corresponding training split, and all normalization parameters are then held fixed during validation and testing.

We instantiate $f_\theta$ as a 10.48-M-parameter joint two-head variant of DigNet++ Large, following the excavation architecture in~\cite{duan2026dignet}. A $3\times3$ convolution maps the six input channels to 32 features. The encoder then produces feature maps of size $50^2\times32$, $25^2\times64$, $13^2\times128$, and $7^2\times256$. Each scale contains three residual blocks, and stride-2 convolutions perform downsampling. Each residual block uses two $3\times3$ convolution--GroupNorm layers, with SiLU activations and a $1\times1$ residual projection when the channel dimension changes.

At the bottleneck, three residual branches with dilation rates 1, 2, and 4 are concatenated and projected from 768 to 256 channels. The projected features are added to the bottleneck input. The terrain decoder bilinearly upsamples the contextual features and concatenates the corresponding encoder features, producing $13^2\times128$, $25^2\times64$, and $50^2\times32$ feature maps. A linear $1\times1$ convolution predicts signed terrain change $\Delta\hat H_t$ without an output activation. In parallel, global average pooling of the contextual bottleneck is followed by a 256-to-128 linear layer, SiLU, a 128-to-1 linear layer, and Softplus to predict nonnegative loaded volume $\hat V_t$. Both heads are trained jointly.

Let $M_p$ denote the valid-pixel mask and let
\begin{equation}
I_p=\mathbf 1\{|\Delta H_p|\geq\tau_H\},
\qquad
w_p=M_p(1+4I_p),
\label{eq:wm-weights}
\end{equation}
where $\tau_H$ is the domain-specific threshold defining the
changed terrain region. Thus, changed pixels receive five times
the weight of other valid pixels. The terrain and loaded-volume losses are
\begin{equation}
\begin{aligned}
\mathcal L_H &=
\frac{\sum_p w_p
\left|\widehat{\widetilde{\Delta H}}_p-
\widetilde{\Delta H}_p\right|}
{\max(\sum_p w_p,1)},\\
\mathcal L_V &=
\frac{\sum_i r_i\,
\operatorname{SL1}_{\beta=1}
(\widehat{\widetilde V}_i-\widetilde V_i)}
{\max(\sum_i r_i,1)},\\
\mathcal L &=\mathcal L_H+\lambda_V\mathcal L_V,
\qquad \lambda_V=1,
\end{aligned}
\label{eq:wm-loss}
\end{equation}
where $r_i$ indicates whether sample $i$ has a valid loaded-volume target. Because both targets are normalized, $\lambda_V=1$ does not imply equal weighting of errors in physical units.

Across all datasets, the domain-specific world models are trained using the AdamW optimizer with a batch size of 32 and an initial learning rate of $10^{-3}$. Sampling is balanced according to the natural grouping units of each dataset, such as excavation episodes, piles, sequences, or physical loading events. Training proceeds for up to 100 epochs, with learning-rate reduction and early stopping based on validation performance.

\subsection{Diffusion Action Proposal}
Given $s_t$, Stage~I constructs an ordered candidate set $\mathcal C_t$ without executing an action. The conditional denoising diffusion model~\cite{ho2020ddpm} is trained exclusively on the MinSlope training split to model the distribution of executed scoop actions. Each transition provides the current terrain state, the executed action, and its observed loaded volume. Training consists of two stages: coverage pretraining, which learns a diverse distribution over the action space, and productive fine-tuning, which biases the proposal distribution toward higher-volume scoops.

To balance action diversity and excavation productivity, the training transitions are divided into five loaded-volume strata ranging from empty to full. Coverage pretraining uses episode-balanced sampling together with inverse-frequency weighting across the five strata, preventing frequently observed intermediate-volume actions from dominating the learned distribution. Productive fine-tuning retains samples from all strata but increases the sampling probability of higher-volume actions. Consequently, the diffusion model preserves multimodal action-space coverage while generating productive scoop candidates more frequently.

The proposal model conditions on normalized terrain height and surface normals together with
\begin{equation}
\mathbf c_t=[\rho_t,g_t]^\top,
\qquad
\rho_t=\frac{V_t^{\rm rem}}{V_0},
\label{eq}
\end{equation}
where $\rho_t$ is the current remaining-volume ratio and $g_t$ is the desired bucket-fill fraction. During training, $g_t$ is the observed fill fraction $V_t/C$. During evaluation, it is fixed to $g^\star=0.85$, corresponding to a target loaded volume $V^\star=g^\star C$. This target conditions proposal generation and the geometric heuristic; the final action is not required to have predicted volume exactly equal to $V^\star$.

A convolutional terrain encoder and a time-conditioned multilayer perceptron predict noise in the normalized seven-dimensional action space. Training uses a cosine schedule with $T=50$ diffusion steps and samples the diffusion index uniformly from ${0,\ldots,T-1}$. At inference, each proposal is generated using the complete 50-step ancestral DDPM reverse process from $\mathcal N(\mathbf 0,\mathbf I_7)$. During evaluation, we control sampling randomness using a fixed seed schedule. For episode seed \(\xi_e\), scoop index \(t\), and resampling round \(r\), the sampling seed is set to \(\xi_e + 9176t + 7919r\). This controls both the initial Gaussian noise and the noise injected during reverse diffusion; the same schedule is used by Diffusion-direct and Diffusion-5+WM.

Each sampling round generates 16 proposals in parallel, which are clipped to $[-1,1]^7$ and examined in generation order. A legal proposal $\mathbf a$ is accepted only if
\begin{equation}
\min_{\mathbf u\in\mathcal C_t}
|\mathbf a-\mathbf u|_2\geq\delta_a,
\qquad \delta_a=0.02,
\label{eq}
\end{equation}
where distance is measured in normalized action space. Sampling stops when five legal and distinct proposals have been accepted or after eight rounds, corresponding to at most 128 examined diffusion proposals. If fewer than five proposals are obtained, the incomplete diffusion set is discarded and replaced by the complete deterministic Geometric set.

\subsection{Geometric Candidate Construction and Admissibility}
Both diffusion proposals and Geometric-5 candidates share the same geometric admissibility filter. Let $\mathbf e$, $\mathbf b$, and $\mathbf c$ denote the entry, cutting-start, and cutting-end points. The maximum approach-corridor height $h_{\rm cor}$ is evaluated at 24 uniformly spaced distances between 6 and 11\,m behind $\mathbf e$, with lateral offsets of $\pm1.65$\,m relative to the action heading. Only samples within the workspace $\Omega=[0,60]^2\,\mathrm m^2$ are considered.

An action is admissible if its seven normalized components are finite and lie in $[-1,1]$, all three points lie within $\Omega$, $h_{\rm cor}\le0.25\,\mathrm m$, and its capacity-limited swept volume $\bar G(H_t,\mathbf a)=\min\{G(H_t,\mathbf a),C\}$ is at least $0.05\,\mathrm{m^3}$. Entry-point displacement must not exceed 9\,m, except for the first scoop or after two consecutive scoops with loaded volume below $0.25\,\mathrm{m^3}$. These conditions define $\mathcal A_{\rm legal}(s_t)$. Candidates must also satisfy the distinctness threshold $\delta_a$. 

Geometric-5 constructs five deterministic candidates around an anchor action whose capacity-limited swept volume approaches $V^\star=g^\star C$. The role targets comprise the anchor, shorter/shallower and longer/deeper cuts, and left/right work-face shifts. In normalized action space, the cut variants jointly offset $\ell$ and $d^{\max}$ by $\pm0.16$, while the lateral variants jointly offset $\psi$ and $y$ by $\pm(0.035,0.080)$. All coordinates are clipped to $[-1,1]$.

An anchor-centered pool of 128 proposals is filtered with the above rules. For each role target $\mathbf r_k$, an unused admissible candidate satisfying the distinctness threshold is selected by minimizing
\begin{equation}
J_{\rm role}(\mathbf a,\mathbf r_k)=
\sum_{j=0}^{6}
\left(\frac{a_j-r_{k,j}}{\sigma_j}\right)^2
+0.02J_{\rm geom}(\mathbf a),
\label{eq:role-score}
\end{equation}
where $\boldsymbol{\sigma}=(0.05,0.10,0.10,0.18,0.18,0.10,1.0)$. The geometric score is
\begin{equation}
\begin{aligned}
J_{\rm geom}(\mathbf a)=&
\left|\bar G(H_t,\mathbf a)-V^\star\right|
+18[h_{\rm cor}-0.05]_+\\
&+0.05d_e+1.5[d_e-7.5]_+,
\end{aligned}
\label{eq:geometric-score}
\end{equation}
where $[x]_+=\max(x,0)$ and $d_e$ is the displacement from the previous entry point, set to zero for the first scoop. Lengths and volumes are expressed in metres and cubic metres.

Geometric-heuristic executes the candidate minimizing $J_{\rm geom}$, whereas Geometric-5+WM evaluates the identical candidate set with the frozen world model and selects the largest predicted loaded volume. This comparison isolates the effect of consequence-based selection.

For WAM, the frozen world model evaluates the five admitted candidates in a single batch, and the action with the highest predicted loaded volume is executed. 

\section{Experimental Setup}
\subsection{Datasets}
For each simulation and real-loader domain, the training- and test-set sample counts are summarized in Table~\ref{tab:multisim}.

\subsubsection{Simulation Datasets}
The four simulation datasets were selected to span complementary task scales, physical modeling assumptions, and data-generation pipelines rather than to compare simulator fidelity. \textbf{MinSlope} provides the large-scale controlled environment used for predictor ablations and continuous-depletion evaluation, with diverse pile geometries and material conditions. \textbf{Chrono}~\cite{fang2021chronogpu} introduces contact-rich granular dynamics through a discrete-element formulation. \textbf{Isaac Sim}~\cite{nvidia2026isaacsim} provides a GPU-accelerated robotic simulation pipeline with a distinct particle and sensing stack. \textbf{Newton MPM}~\cite{newton2025} complements the particle-based settings with a material-point formulation for large terrain deformation. Together, these datasets test whether the same action-conditioned prediction interface can be learned across different physical solvers, response scales, and data regimes. 

\subsubsection{Real-Loader Dataset}
\begin{figure}[t]
\centering
\includegraphics[width=\columnwidth]{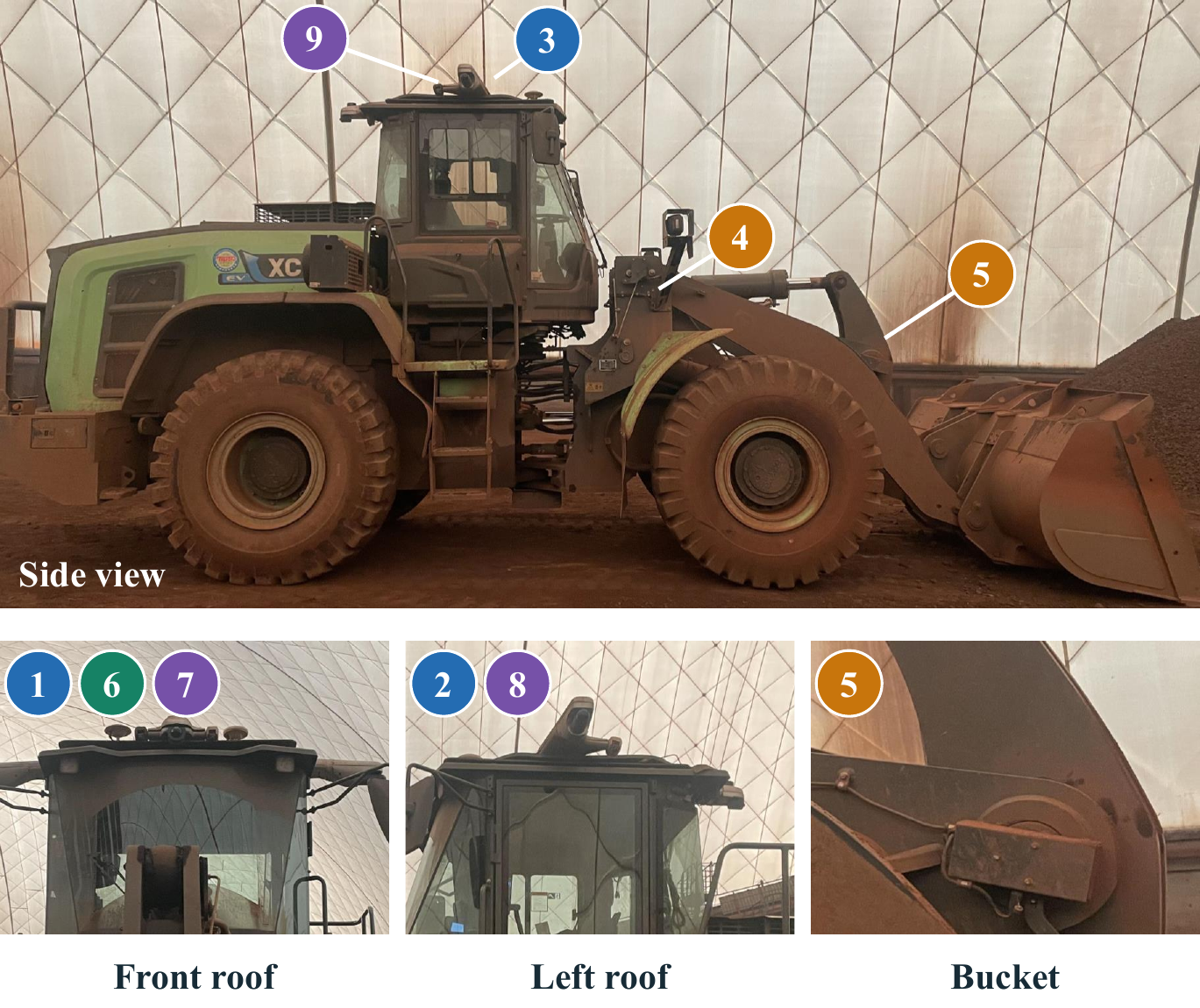}
\caption{Real-loader platform and sensor configuration. LiDARs (1–3) and fisheye cameras (7–9) provide surround-view pile observations. Boom and bucket angle sensors (4 and 5, respectively) measure articulation angles, while the front-facing pinhole camera (6) supports image-based loaded-volume estimation.}
\label{fig:real_loader_site}
\end{figure}

The real-world dataset was collected on a physical wheel loader platform at the site shown in Fig.~\ref{fig:real_loader_site}. The platform is an XCMG XC958EV electric wheel loader with a rated payload of 5.5 t, deployed within a port for iron ore powder pile transfer operations. An RTK receiver is mounted on top of the cabin to provide ego-pose estimation. Three LiDARs and three fisheye cameras are mounted at the front, left, and right sides of the cabin, providing aggregated pile point clouds for heightmap computation. Both the RTK and LiDAR frames are referenced to the center of the rear axle. Two angle sensors are mounted on the boom and bucket for bucket pose estimation, and a front-facing pinhole camera is installed to observe the bucket and the heaped ore above its rim.

All data were collected from the wheel loader during routine port production without interrupting operations. Each sample pairs an input (pre-scooping heightmap and bucket trajectory) with an output (post-scooping heightmap and scooped volume).

Surround-view image semantics, LiDAR scans, and ego poses are fused into a semantic occupancy grid for heightmap extraction, with a normal map derived from height differences. Bucket tip poses from two angle sensors define the loading and exit points, around which trajectory endpoints are sampled. Two action-derived scalar fields are further computed from the trajectory. The sweep mask projects the complete bucket trajectory onto the 0.2 m-resolution terrain grid, with adjacent trajectory frames linearly interpolated at 0.1 m intervals; grids traversed by the bucket's 2D projection contour are marked 1 and the rest 0, identifying the effective interaction region. For each swept grid, the dig depth is computed as the difference between the initial terrain height and the lowest height of the bucket cutting edge over the entire trajectory, truncated at zero, representing the maximum theoretical penetration depth at that location. The post-scooping cloud is ICP-aligned to the pre-scooping one in a unified tip-origin frame.

The scooped volume is estimated from front-view images captured at a fixed retreat position after each scoop. A heap segmentation model is trained to obtain the trapezoid-fitted heap, whose contour is fitted with a trapezoid and then reconstructed as a hip-roof-shaped pentahedron, scaled by the known bucket width and the endpoint pixel distance, provided by a bucket endpoint detector. The metric heaped volume is then calculated and added to the struck capacity to yield the estimated scooped volume.

\subsection{SAC Baseline, Metrics, and Evaluation Control}
We use soft actor--critic (SAC)~\cite{haarnoja2018soft} as a system-level competitive baseline for the complete WAM policy. Let $E_t$ denote drive energy, $\theta_t$ and $\phi_t$ the maximum absolute pitch and roll in degrees, $u_t^{\rm slip}$ the maximum slip, and $[x]_+=\max\{0,x\}$. The per-scoop reward is
\begin{equation}
\begin{aligned}
r_t^{\rm dyn}={}&V_t-0.10u_t^{\rm slip}
-2\!\times\!10^{-8}E_t\\
&-8\!\times\!10^{-4}
\left([\theta_t-12]_+^2+[\phi_t-10]_+^2\right),\\
r_t={}&\frac{V_t}{C}-0.08
-0.15\mathbf 1\{V_t<0.25\}\\
&+2\mathbf 1\{\rho_{t+1}<0.10\}
+0.10\frac{r_t^{\rm dyn}}{C}.
\end{aligned}
\label{eq:sacreward}
\end{equation}

The loaded-volume term rewards excavation progress, while the action cost and low-volume penalty discourage inefficient scoops. Reaching the 10\% residual threshold provides a terminal bonus. SAC receives a 231-dimensional observation comprising a $15\times15$ downsampled height map and six context variables: pile-peak coordinates and height, bulk density, cohesion, and the remaining-volume ratio. All inputs are clipped to their physical ranges and scaled to $[-1,1]$. Internal friction, bucket-wall friction, and velocity drag affect the simulator dynamics but are not policy inputs.

The actor and each critic use two 256-unit hidden layers. SAC is
trained for 300{,}000 environment interactions using a learning rate
of $3\times10^{-4}$, batch size 256, discount factor $0.995$, target
smoothing coefficient $0.005$, a 250{,}000-transition replay buffer,
and 20{,}000 warm-up steps, with one gradient update per environment
step. Training samples resets from 24 geometries and four material
settings, and evaluation uses the deterministic SAC-300k checkpoint.

SAC and WAM share the normalized action representation, its mapping to physical quantities, and the MinSlope execution environment. The experiment-specific environment sets the termination threshold to 10\% residual volume and limits each episode to 1600 scoops. An episode succeeds when its residual pile volume is no more than 10\% of the initial volume. For a successful episode, $N_{10}\in\mathbb N$ denotes the number of executed scoops required to reach this threshold. Episodes that do not reach the threshold within 1600 scoops are failures and are treated as right-censored rather than assigned $N_{10}=1600$.

The test-set comparison separates proposal generation from candidate selection. Geometric-heuristic constructs five fixed-role geometric candidates and executes the candidate with the smallest handcrafted heuristic score. Geometric-5+WM uses the same five candidates but executes the one with the largest world-model-predicted loaded volume. Diffusion-direct and Diffusion-5+WM use the same frozen DDPM, seed schedule, geometric admissibility filter, resampling budget, fallback rule, and candidate ordering. Diffusion-direct executes the first candidate \(D_0\) in the resulting ordered set without consulting the world model, whereas Diffusion-5+WM selects the candidate with the largest predicted loaded volume. The comparison is paired by initial episode configuration and seed, with each policy running independently in closed loop. After their selected actions diverge, subsequent terrain states and candidate sets may differ. Thus, this ablation changes the selection rule while keeping the proposal-generation procedure fixed.

Prediction quality is reported as full-ROI terrain MAE, changed-region MAE and IoU, and loaded-volume MAE. Closed-loop performance is measured by success at 10\% residual volume, $N_{10}$, and mean loaded volume.

\section{Results}
We organize all results around one question: does predicting candidate consequences before execution improve continuous-depletion decisions? We first compare Diffusion-direct with WAM under the same proposal mechanism, then report the complementary geometric comparison and the complete-system comparison with SAC. Prediction accuracy, architecture, runtime, and additional data pipelines are supporting diagnostics.

\subsection{From Candidate Prediction to Closed-Loop Decisions}
The test-set comparisons for both proposal mechanisms are summarized in Table~\ref{tab:n10}. The most direct test of decision value holds the proposal mechanism fixed and changes only how one candidate is selected. On the fixed 32-episode test set, Diffusion-direct executes the first accepted member of the Diffusion-5 set, whereas WAM evaluates its generated candidates before execution and selects the candidate with the largest predicted loaded volume. Both policies complete 32/32 episodes, but world-model ranking reduces mean $N_{10}$ from 651.8 to 540.6 scoops, improves all 32 paired episodes, and raises mean loaded volume from 4.03 to 4.78\,m$^3$. This matched-candidate comparison shows that the gain is not explained by diffusion proposal generation alone: consequence prediction changes which candidate is executed and improves the resulting depletion process.

The geometric comparison provides complementary evidence. Replacing the handcrafted selection rule with world-model ranking increases the number of completed episodes from 15/32 to 22/32 and reduces the mean terminal residual volume across all episodes from 22.61\% to 16.80\%.

\begin{table}[t]
\caption{Continuous-depletion ablation on 32 MinSlope test episodes. Mean $N_{10}$ uses successful episodes; residual and load means use all episodes.}
\label{tab:n10}
\centering
\footnotesize
\setlength{\tabcolsep}{2.2pt}
\begin{tabular}{@{}lcccc@{}}
\toprule
\multirow{2}{*}{Policy}
& \multirow{2}{*}{Success}
& Mean
& Terminal
& Mean load\\
&
& $N_{10}$
& resid. (\%)
& (m$^3$)\\
\midrule
Diffusion-direct
& 32/32
& 651.8
& 9.90
& 4.03\\

\textbf{Diffusion-5+WM}
& \textbf{32/32}
& \textbf{540.6}
& 9.92
& \textbf{4.78}\\
\midrule

Geometric-heuristic
& 15/32
& 740.1
& 22.61
& 2.36\\

Geometric-5+WM
& 22/32
& 549.2
& 16.80
& 3.66\\
\midrule

SAC
& 29/32
& 596.2
& 11.00
& 4.19\\
\bottomrule
\end{tabular}
\end{table}

\subsection{Supporting Prediction Diagnostics}
Table~\ref{tab:prediction} compares three input spatial extents: ROI-5 (5 × 5 m), ROI-10 (10 × 10 m), and Global-60 (60 × 60 m). The local ROI coordinate origin is defined by the midpoint of the two lower bucket keypoints at the start of the candidate action. Global-60 uses a fixed world-coordinate view covering the entire pile. All three configurations use the same spatial resolution of 0.2 m per cell. Compared with ROI-5, ROI-10 reduces changed-region MAE and loaded-volume MAE by approximately 43\% and 34\% respectively, and increases IoU from 0.87 to 0.93, while increasing median batch-1 inference latency from 3.33 to 3.65 ms. Global-60 provides no accuracy improvement over ROI-10 and requires 23.73 ms per inference. We therefore adopt ROI-10.

\begin{figure*}[t]
\centering
\includegraphics[width=\textwidth]{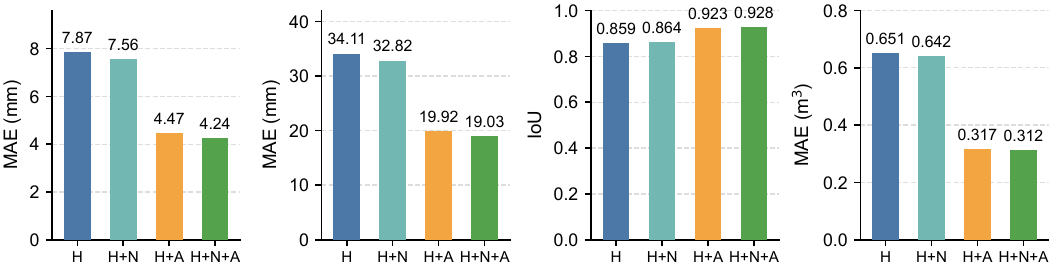}
\par\smallskip
\makebox[\textwidth][c]{%
\parbox{0.235\textwidth}{\centering\footnotesize (a) Terrain MAE}\hfill
\parbox{0.235\textwidth}{\centering\footnotesize (b) Changed-region MAE}\hfill
\parbox{0.235\textwidth}{\centering\footnotesize (c) IoU}\hfill
\parbox{0.235\textwidth}{\centering\footnotesize (d) Loaded-volume MAE}}
\caption{Input ablation at fixed ROI-10. H denotes the terrain height map, N the three surface-normal channels, and A the action-derived sweep mask and dig-depth map.}
\label{fig:input}
\end{figure*}

Figure~\ref{fig:input} fixes ROI-10 and varies only the input channels on the 25,349-sample validation set. Adding the action maps to the terrain height map reduces the terrain, changed-region, and loaded-volume MAEs by 43.20\%, 41.60\%, and 51.31\%, respectively. Surface normals alone provide only modest gains. Once the action maps are included, further adding surface normals reduces the terrain and changed-region MAEs by 5.15\% and 4.47\%, increases IoU by 0.005, and reduces the loaded-volume MAE by 1.58\%. These results indicate that the action-derived sweep and dig-depth maps provide the dominant information for predicting excavation outcomes.

\begin{table}[t]
\caption{Spatial-support ablation on MinSlope validation set.
Latency is the median FP32 batch-1 forward time.}
\label{tab:prediction}
\centering
\footnotesize
\setlength{\tabcolsep}{3pt}
\begin{tabular}{@{}lcccc@{}}
\toprule
\multirow{2}{*}{Scope}
& Change MAE & IoU & Volume MAE & $t_1$ \\
& (mm)$\downarrow$ & $\uparrow$
& (m$^3$)$\downarrow$ & (ms)$\downarrow$ \\
\midrule
ROI-5
& 33.41 & 0.87 & 0.47 & \textbf{3.33} \\
ROI-10
& \textbf{19.03} & \textbf{0.93}
& \textbf{0.31} & 3.65 \\
Global-60
& 23.74 & 0.87 & 0.36 & 23.73 \\
\bottomrule
\end{tabular}
\end{table}

Table~\ref{tab:arch} compares the prediction accuracy, model size, and inference efficiency of U-Net~\cite{ronneberger2015unet}, ResUNet, two DigNet++ variants, and Terrain-JEPA. The DigNet++ variants follow the excavation architecture in~\cite{duan2026dignet}, whereas Terrain-JEPA adapts the latent feature-prediction principle of I-JEPA and V-JEPA~\cite{assran2023ijepa,bardes2024vjepa} to action-conditioned height-map transitions. All models are profiled on an NVIDIA GeForce RTX 3070 Laptop GPU using the same $[B,6,50,50]$ ROI-10 input, and the reported latencies are model-only FP32 medians for a batch of five candidate actions.

DigNet++ Large achieves the highest IoU of 0.928 while maintaining a terrain MAE of 4.237\,mm. Terrain-JEPA reduces terrain MAE by only 0.20\%, but uses 23.2\% more parameters, requires 10.4\% longer inference time, and produces a slightly lower IoU. U-Net uses only 18.9\% of the parameters of DigNet++ Large and is 3.01$\times$ faster for a batch of five candidates, but its terrain MAE is 15.4\% higher and its IoU is lower by 0.019. DigNet++ Large evaluates all five candidate actions in 4.280\,ms, corresponding to 0.856\,ms per candidate. Considering prediction accuracy, model size, and batch inference latency together, we select DigNet++ Large as the world-model backbone.

\begin{table}[t]
\caption{World-model accuracy and inference efficiency on the MinSlope validation set.}
\label{tab:arch}
\centering
\footnotesize
\setlength{\tabcolsep}{2.2pt}
\begin{tabular}{@{}lcccc@{}}
\toprule
\multirow{2}{*}{Model}
& Terrain MAE
& \multirow{2}{*}{IoU$\uparrow$}
& Parameters
& Batch-5 latency\\
& (mm)$\downarrow$
&
& (M)$\downarrow$
& (ms)$\downarrow$\\
\midrule
U-Net        & 4.891 & 0.909          & \textbf{1.98} & \textbf{1.424}\\
ResUNet      & 5.021 & 0.905          & 2.82           & 1.744\\
DigNet++ S   & 4.589 & 0.914          & 2.11           & 2.209\\
DigNet++ L   & 4.237 & \textbf{0.928} & 10.48          & 4.280\\
Terrain-JEPA & \textbf{4.229} & 0.926 & 12.91          & 4.725\\
\bottomrule
\end{tabular}
\end{table}

\subsection{Prediction Interface on Additional Data Pipelines}
Table~\ref{tab:multisim} evaluates the same action-conditioned prediction interface on four simulation datasets and the real-loader dataset. A separate DigNet++ Large model is trained for each domain using common terrain- and action-derived inputs. Despite differences in physical formulation, response scale, and data regime, the changed-region IoU ranges from 0.641 to 0.928 across the four simulation domains, indicating that the interface can capture action-induced terrain responses under different simulation mechanisms. 

On the real-loader dataset, the model achieves terrain and changed-region MAEs of 0.054 m and 0.071 m, a changed-region IoU of 0.640, and a loaded-volume MAE of 0.148 m$^3$. These results show that the same prediction interface can be learned from onboard measurements despite sensor noise, point-cloud registration errors, and image-derived loaded-volume labels. Overall, the results establish the domain-specific learnability of the proposed interface in both simulation and real-world operation.

\begin{table}[t]
\caption{Domain-specific evaluation of DigNet++ Large.}
\label{tab:multisim}
\centering
\setlength{\tabcolsep}{3.0pt}
\begin{tabular}{@{}lrrrrr@{}}
\toprule
\multirow{2}{*}{Domain}
& Train
& Test
& Terrain
& \multirow{2}{*}{IoU$\uparrow$}
& Volume\\
& $N$
& $N$
& MAE (m)$\downarrow$
&
& MAE (m$^3$)$\downarrow$\\
\midrule
MinSlope  & 72{,}241 & 25{,}014 & 0.004 & 0.928 & 0.312\\
\midrule
Chrono    & 506       & 127      & 0.048 & 0.695 & 0.218\\
Isaac Sim & 1{,}672   & 188      & 0.004 & 0.692 & 0.066\\
Newton MPM & 662      & 142      & 0.001 & 0.641 & 0.003\\
\midrule
Real loader & 1{,}015 & 130      & 0.054 & 0.640 & 0.148\\
\bottomrule
\end{tabular}
\end{table}

\subsection{Onboard Deployment}
\begin{figure}[t]
\centering
\includegraphics[width=\columnwidth]{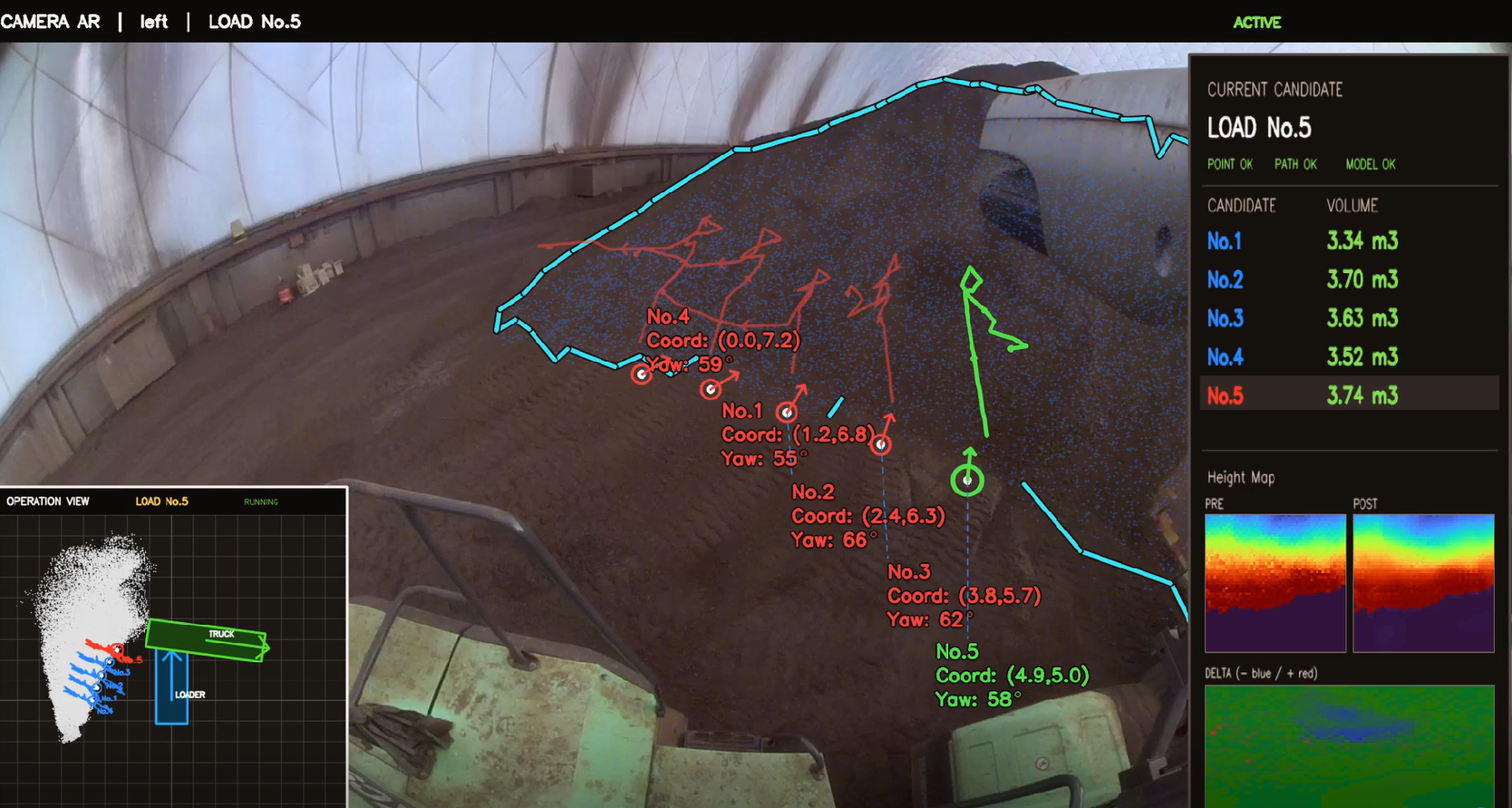}
\caption{Onboard visualization during real-loader deployment. Five candidate scoop trajectories are overlaid on the camera view, with the selected candidate highlighted in green. The right panel displays candidate volume predictions and terrain-map visualizations.}
\label{fig:onboard_deployment}
\end{figure}

The DigNet++ Large world-model was deployed in a ROS2 node on an NVIDIA Jetson AGX Orin equipped with 64 GB of memory. Point cloud keyframes are extracted from recent LiDAR scans and concatenated to construct the aggregated point cloud for online heightmap generation. The bucket tip trajectory is also generated and validated in real time and fed into the model together with the heightmap. Online inference was performed using a mixed-precision C++/TensorRT pipeline with a batch size of 5, and the predicted height change and volume were then denormalized back to physical units, followed by sanity checks. For each batch of 5 candidate trajectories, preprocessing, inference, and postprocessing took 60.6, 4.1, and 7.7 ms, respectively, for a total latency of 72.4 ms. Figure~\ref {fig:onboard_deployment} shows the onboard visualization of candidate actions and model outputs during real loader deployment.

\section{Conclusion and Limitations}
We introduced continuous pile depletion to 10\% residual volume to evaluate the decision value of world-model predictions. WAM uses diffusion proposals, geometric admission, and frozen terrain--volume prediction to rank scoops before execution. On geometry-disjoint MinSlope tests, predicted-volume ranking reduces mean scoop count from 651.8 to 540.6 (17.1\%), preserves 32/32 completion, and improves all paired episodes. System-level comparison shows WAM completes 32/32 episodes versus 29/32 for SAC. Ablations identify action-derived sweep and dig-depth maps as the main source of prediction improvement. Domain-specific models show that the prediction interface is learnable across four simulation datasets and event-disjoint real-loader data. Together, these results show that pre-execution consequence prediction improves continuous-depletion action selection.

Quantitative gains are established in MinSlope, whose height-field dynamics simplify granular behavior. Although WAM operated autonomously on a full-size loader, the limited trials demonstrate closed-loop feasibility, not statistically supported efficiency gains. Predictors are trained separately by domain, precluding claims of zero-shot transfer. The selector ranks immediate predicted load without explicit look-ahead over terrain change. Future work will conduct repeated field trials across pile geometries and materials for statistically powered comparisons and develop terrain-aware multi-step objectives.

\bibliographystyle{IEEEtran}
\bibliography{references}
\end{document}